# Learning to Rank With Bregman Divergences and Monotone Retargeting

**Sreangsu Acharyya** [*]
Dept. Electrical Engineering
University of Texas Austin

**Oluwasanmi Koyejo**[*]
Dept. Electrical Engineering
University of Texas Austin

**Joydeep Ghosh**
Dept. Electrical Engineering
University of Texas Austin

## Abstract

This paper introduces a novel approach for learning to rank (LETOR) based on the notion of monotone retargeting. It involves minimizing a divergence between all monotonic increasing transformations of the training scores and a parameterized prediction function. The minimization is both over the transformations as well as over the parameters. It is applied to Bregman divergences, a large class of "distance like" functions that were recently shown to be the unique class that is statistically consistent with the normalized discounted gain (NDCG) criterion [19]. The algorithm uses alternating projection style updates, in which one set of simultaneous projections can be computed independent of the Bregman divergence and the other reduces to parameter estimation of a generalized linear model. This results in easily implemented, efficiently parallelizable algorithm for the LETOR task that enjoys global optimum guarantees under mild conditions. We present empirical results on benchmark datasets showing that this approach can outperform the state of the art NDCG consistent techniques.

## 1 Introduction

Structured output space models [1] have dominated the task of learning to rank (LETOR). Regression based models have been justifiably superseded by pairwise models [11], which in turn are being gradually displaced by list-wise approaches [6, 17]. This trend has on one hand greatly improved the quality of the predictions obtained but on the other hand has come at the cost of additional complexity and computation. The cost functions of structured models are often defined directly on the combinatorial space of permutations, which significantly increase the difficulty of learning and optimization compared to regression based approaches. We propose an approach to the LETOR task that retains the simplicity of the regression based models, is simple to implement, is embarrassingly parallelizable, and yet is a function of ordering alone. Furthermore, MR enjoys strong guarantees of convergence, statistical consistency under uncertainty and a global minimum under mild conditions. Our experiments on benchmark datasets show that the proposed approach outperforms state of the art models in terms of several common LETOR metrics.

We adapt regression to the LETOR task by using Bregman divergences and monotone retargeting (MR). MR is a novel technique that we introduce in the paper and Bregman divergences [5] are a family of "distance like" functions well studied in optimization [8], statistics and machine learning [2] due to their one to one connections with modeling uncertainty using exponential family distributions. Bregman divergences are the unique class of strongly statistically consistent surrogate cost functions for the NDCG criterion [19], a de facto standard of ranking quality. In addition to these statistical properties, Bregman divergences have several properties useful for optimization and specifically useful for ranking. The LETOR task decomposes into subproblems that are equivalent to estimating (unconstrained as well as constrained) generalized linear models. The Bregman divergence machinery provides easy to implement, scalable algorithms for them, with a user chosen level of granularity of parallelism. We hope the reader will appreciate the flexibility of choosing an appropriate divergence to encode desirable properties on the rankings while enjoying the strong guarantees that come with the family.

We introduce MR by first discussing direct regression of rank scores and highlighting its primary deficiency: its attempt to fit the scores exactly. An exact fit is unnecessary since any score that induces the correct ordering is sufficient. MR addresses this problem by searching for a order preserving transformation of the target scores that may be easier for the regressor to fit: hence the name "retargeting".

Let us briefly sketch our line of attack. In section 2 we present a method to reduce the optimization over the infinite class of all monotonic increasing functions to that of alternating projection over a finite dimensional vector space. In section 3.2.3 we show when that optimization problem is jointly convex by resolving the question of joint convexity of the Fenchel-Young gap. This result is important in its own right. We introduce Bregman divergences in section 3 and discuss properties that make them particularly suited to the ranking task. We show (i) that one set of the alternating projections can be computed in a Bregman divergence independent fashion 3.2.1, and (ii) separable Bregman divergences allow us to use sorting 3.2.2 that would have otherwise required exhaustive combinatorial enumeration or solving a linear assignment problem repeatedly.

---
[*] Both authors contributed equally.

**Notation:** Vectors are denoted by bold lower case letters, matrices are capitalized. $x^\dagger$ denotes the transpose of the vector $x$, $||x||$ denotes the $L_2$ norm. $\text{Diag}(x)$ denotes a diagonal matrix with its diagonal set to the vector $x$. $\text{Adj-Diff}(x)$ denotes a vector obtained by taking adjacent difference of consecutive components of $\begin{bmatrix} x \\ 0 \end{bmatrix}$. Thus $\text{Cum-Sum}(\text{Adj-Diff}(x)) = x$. A vector $x$ is defined to be in *descending order* if $x_i \geq x_j$ if $i > j$, the set of such vectors is denoted by $\mathcal{R}\downarrow$. Vector $x$ is isotonic with $y$ if $x_i \geq x_j$ then $y_i \geq y_j$. The unit simplex is denoted by $\Delta$ and the positive orthant by $\mathbb{R}_+^d$. $\psi(\cdot)$ is used to denote the Legendre dual of the function $\phi(\cdot)$. Partitions of sets are denoted by $\Pi$ and $P$.

## 2 Monotone Retargeting

We introduce our formulation of learning to rank, this consists of a set of queries $\mathcal{Q} = \{q_1, q_i \ldots q_{|\mathcal{Q}|}\}$ and a set of items $\mathcal{V}$ that are to be ranked in the context of the queries. For every query $q_i$, there is a subset $\mathcal{V}_i \subset \mathcal{V}$ whose elements have been ordered, based on their relevance to the query. This ordering is customarily expressed via a rank score vector $\tilde{r}_i \in \mathbb{R}^{d_i=|\mathcal{V}_i|}$ whose components $\tilde{r}_{ij}$ correspond to items in $\mathcal{V}_i$. Beyond establishing an order over the set $\mathcal{V}_i$, the actual values of $\tilde{r}_{ij}$ are of no significance. For a query $q_i$ the index $j$ of $\tilde{r}_{ij}$ is local to the set $\mathcal{V}_i$ hence $\tilde{r}_{ij}$ and $\tilde{r}_{kj}$ need not correspond to the same object. We shall further assume, with no loss in generality, that the subscript $j$ is assigned such that $\tilde{r}_{ij}$ is in a descending order for any $\mathcal{V}_i$. Note that $\tilde{r}_i$ induces a partial order if the number of unique values $k_i$ in the vector is less than $d_i$.

For every query-object pair $\{q_i, v_{ij}\}$ a feature vector $\mathbb{R}^n \ni a_{ij} = F(q_i, v_{ij})$ is pre-computed. The subset of training data pertinent to any query $q_i$ is the pair $\{\tilde{r}_i, A_i\}$ and is called its qset. Thus, the column vector $\tilde{r}_i$ consists of the rank-scores $\tilde{r}_{ij}$ and $A_i$ is a matrix whose $j^{th}$ row is $a_{ij}^\dagger$.

Given a loss function $D_i : \mathbb{R}^{|\mathcal{V}_i|} \times \mathbb{R}^{|\mathcal{V}_i|} \mapsto \mathbb{R}_+$ we may define the regression problem $\min_{w} \sum_i D(\tilde{r}_i, f(A_i, w))$ where $f : \mathbb{R}^{|\mathcal{V}_i| \times n} \times \mathbb{R}^n \mapsto \mathbb{R}^{|\mathcal{V}_i|}$ is some fixed parametric form with the parameter $w$. As discussed, this is unnecessarily stringent for ranking. A better alternative is:

$$\min_{w, \Upsilon_i \in \mathcal{M}} \sum_i D_i\big(\tilde{r}_i, \Upsilon_i \circ f(A_i, w)\big),$$

where $\Upsilon_i : \mathbb{R}^{|\mathcal{V}_i|} \mapsto \mathbb{R}^{|\mathcal{V}_i|}$ transforms the component of its argument by a fixed monotonic increasing function $\Upsilon_i$, and $\mathcal{M}$ is the class of all such functions. Now $f(A_i, w)$ no longer need to equal $\tilde{r}_i$ point-wise to incur zero loss. It is sufficient for some monotonic increasing transform of $f(A_i, w)$ to do so. With no loss in generality of modeling, we may apply the monotonic transform to $\tilde{r}_i$ instead. This avoids the minimization over the function composition, but the need for minimizing over the set of all monotone functions remains. One possible way to eliminate the minimization over the function space is to restrict our attention to some parametric family in $\mathcal{M}$ at the expense of generality. Instead, with no loss in generality, the optimization over the infinite space of functions $\mathcal{M}$ can be converted into one over finite dimensional vector spaces $\mathbb{R}^{|\mathcal{V}_j|}$, provided we have a finite characterization of the constraint set $\mathcal{R}\downarrow_i$:

$$\min_{w, r \in \mathcal{R}\downarrow_i} \sum_i D_i(r_i, f(A_i, w)) \text{ s.t. } \mathcal{R}\downarrow_i = \{r \mid_{M(\tilde{r}_i)=r}^{\exists M \in \mathcal{M}}\}. \quad (1)$$

**The Set $\mathcal{R}\downarrow_i$:** The convex composition $r = \alpha r_1 + (1-\alpha) r_2$ of two isotonic vectors $r_1$ and $r_2$ preserves isotonicity, as does the scaling $\alpha r_1$ for any $\alpha \in \mathbb{R}_+$. Hence the set $\mathcal{R}\downarrow_i$ is a convex cone. This makes the problem computationally tractable because the set can be described entirely by its extreme rays, or by the extreme rays of its polar. We claim the set $\mathcal{R}\downarrow_i$ can be expressed as the image of the set $\{\mathbb{R}_+\}^{|\mathcal{V}_i|-1} \times \mathbb{R}$ under a linear transformation by a particular upper triangular matrix $U$ with positive entries:

$$\mathcal{R}\downarrow_i = Ux \quad \text{s.t.} \quad x \in \{\mathbb{R}_+\}^{|\mathcal{V}_i|-1} \times \mathbb{R}$$

The matrix $U$ is not unique and can be generated from any vector $v \in \mathbb{R}_+^{|\mathcal{V}_i|}$, but as we shall see, any member from the allowed class of $U$ is sufficient for a *exhaustive* representation of $\mathcal{R}\downarrow_i$.

**Lemma 1.** *The set of all vectors in $\mathbb{R}^d$ that are sorted in a descending order is given by $Ux$ s.t. $x \in \{\mathbb{R}_+\}^{|\mathcal{V}_i|-1} \times \mathbb{R}$ where $U$ is a triangular matrix generated from a vector $v \in \mathbb{R}_+^d$ such that the $i^{th}$ row $U(i,:)$ is $\{0\}^{i-1} \times v(i:)$*

*Proof.* Consider solving $Ux = \tilde{r}_i$ for any vector $\tilde{r}_i$ sorted in descending order. We have $x = (\text{Diag})^{-1}(v) \times \text{Adj-Diff}(\tilde{r}_i)$ which is in $\{\mathbb{R}_+\}^{|\mathcal{V}_i|-1} \times \mathbb{R}$  □

For regression functions capable of fitting an arbitrary additive offset, no generality is lost by constraining the last component to be non-negative.

In addition to the set $\mathcal{R}\downarrow_i$ we shall make frequent use of the set of all discrete probability distributions that are in descending order, i.e. $\mathcal{R}\downarrow_i \cap \Delta_i$ that we represent by $\Delta_o^i$.

We give a similar representation of this set by generating an upper triangular matrix $T$ from the vector $\boldsymbol{v}_\Delta = \{1, \frac{1}{2}, \cdots \frac{1}{i} \cdots \frac{1}{d}\}$ and considering $\boldsymbol{x} \in \Delta$.

**Lemma 2.** *The set $\Delta_o$ of all discrete probability distributions of dimension $d$ that are in descending order is the image $T\boldsymbol{x}$ s.t. $\boldsymbol{x} \in \Delta$ where $T$ is an upper triangular matrix generated from the vector $\boldsymbol{v}_\Delta = \{1, \frac{1}{2} \cdots \frac{1}{d}\}$ such that $T(i,:) = \{0\}^{i-1} \times \boldsymbol{v}_\Delta(i:)$*

*Proof.* The proof follows Lemma (1). $T\boldsymbol{x}$ is in the simplex $\Delta$ because it is a convex combination of vectors in $\Delta$. □

With appropriate choices of the distance like function $D_i(\cdot, \cdot)$ and the curve fitting function $f(\cdot, \cdot)$ we can transform (1) into a bi-convex optimization [1] problem over a product of convex sets. We choose $D_i(\cdot, \cdot)$ to be a Bregman divergence $D_\phi(\cdot||\cdot)$, defined in Section 3.1, and $f(\boldsymbol{A}_i, \boldsymbol{w})$ to be $(\nabla \phi)^{-1}(\langle \boldsymbol{A}_i, \boldsymbol{w}\rangle)$, leading to the formulation:

$$\min_{\boldsymbol{w} \in \mathcal{W}, \boldsymbol{r} \in \mathcal{R}\downarrow_i} \sum_{i=1}^{|\mathcal{Q}|} \frac{1}{|\mathcal{V}_i|} D_\phi\big(\boldsymbol{r}_i \big|\big| (\nabla\phi)^{-1}(\langle \boldsymbol{A}_i, \boldsymbol{w}\rangle)\big). \quad (2)$$

Coordinate-wise updates of (2) are equivalent to learning canonical GLMs under linear constraints for which scalable techniques are known [9]. The LETOR task has additional structure that allows more efficient solutions.

## 3 Background

We make heavy use of identities and algorithms associated with Bregman divergences, some that to the best of our knowledge are new e.g. Theorem 2, Lemmata 3, 4 and independent proof of Theorem 1. Theorem 2, and Lemmata 3, and 4 are particularly relevant to ranking. The purpose of this section is to collect these results in a single place.

### 3.1 Definitions

**Bregman Divergence:** Let $\boldsymbol{\phi} : \Theta \mapsto \mathbb{R}$, $\Theta = \operatorname{dom} \boldsymbol{\phi} \subseteq \mathbb{R}^d$ be a strictly convex, closed function, differentiable on int $\Theta$. The corresponding Bregman divergence $D_\phi(\cdot||\cdot) : \operatorname{dom}(\boldsymbol{\phi}) \times \operatorname{int}(\operatorname{dom}(\boldsymbol{\phi})) \mapsto \mathbb{R}_+$ is defined as $D_\phi(\boldsymbol{x}||\boldsymbol{y}) \triangleq \phi(\boldsymbol{x}) - \phi(\boldsymbol{y}) - \langle \boldsymbol{x} - \boldsymbol{y}, \nabla \phi(\boldsymbol{y})\rangle$. From strict convexity it follows that $D_\phi(\boldsymbol{x}||\boldsymbol{y}) \geq 0$ and $D_\phi(\boldsymbol{x}||\boldsymbol{y}) = 0$ iff $\boldsymbol{x} = \boldsymbol{y}$. Bregman divergences are (strictly) convex in their first argument, but not necessarily convex in their second.

In this paper we only consider functions of the form $\phi(\cdot) : \mathbb{R}^n \ni \boldsymbol{x} \mapsto \sum_i w_i \phi(x_i)$ that are weighted sums of *identical* scalar convex functions applied to each component. We refer to this class as *weighted, identically separable* (**WIS**) or simply **IS** if the weights are equal. This class has properties particularly suited to ranking. Mahalanobis distance with diagonal $W$, weighted KL divergence $wKL(\boldsymbol{x}||\boldsymbol{y})$ and weighted and shifted generalized I-divergence $wGI(\boldsymbol{x}||\boldsymbol{y})$ are in this family (Table 1).

[1] A biconvex function is a function of two arguments such that with any one of its argument fixed the function is convex in the other argument.

| $\phi(\boldsymbol{x})$ | $D_\phi(\boldsymbol{x}||\boldsymbol{y})$ |
|---|---|
| $\frac{1}{2}\|\|\boldsymbol{x}\|\|_W^2$ | $\frac{1}{2}\|\|\boldsymbol{x}-\boldsymbol{y}\|\|_W^2$ |
| $\sum_i w_i x_i \log x_i \ \boldsymbol{x} \in \Delta$ | $wKL(\boldsymbol{x}||\boldsymbol{y}) = \sum_i w_i x_i \log(\frac{x_i}{y_i})$ |
| $\sum_i w_i(x_i \log x_i - x_i)$ $\boldsymbol{x} \in \mathbb{R}_+^d$ | $wGI(\boldsymbol{x}||\boldsymbol{y})$ $= \sum_i w_i\big((x_i-1)\log(\frac{x_i-1}{y_i-1}) - x_i + y_i\big)$ |

Table 1: Examples of WIS Bregman divergences.

**Bregman Projection:** Given a closed convex set $\mathcal{S}$, the Bregman-projection of $\boldsymbol{q}$ on $\mathcal{S}$ is $\operatorname{Proj}^\phi(q, \mathcal{S}) \triangleq \operatorname{Argmin}_{\boldsymbol{p}} D_\phi(\boldsymbol{p}||\boldsymbol{q}) \ \boldsymbol{p} \in S$.

A function $\phi(\cdot)$ has **modulus of strong convexity** $s$ if $\phi(\alpha\boldsymbol{x} + (1-\alpha)\boldsymbol{y}) \leq \alpha\phi(\boldsymbol{x}) + (1-\alpha)\phi(\boldsymbol{y}) - \frac{s}{2}\alpha(1-\alpha)\|\|\boldsymbol{x}-\boldsymbol{y}\|\|^2$. For a twice differentiable $\phi(\boldsymbol{x})$ this means that eigenvalues of its Hessian are lower bounded by $s$.

The **Legendre conjugate** $\psi(\cdot)$ of the function $\phi(\cdot)$ is defined as $(\phi)^*(\boldsymbol{x}) \triangleq \psi(\boldsymbol{x}) \triangleq \sup_{\boldsymbol{\lambda}}(\langle \boldsymbol{\lambda}, \boldsymbol{x}\rangle - \phi(\boldsymbol{\lambda}))$. If $\phi(\cdot)$ is a convex function of Legendre type [21], as will always be the case in this paper, $\big((\phi)^*\big)^*(\cdot) = \phi(\cdot)$ and $(\nabla\phi(\cdot))^{-1} = \nabla\psi(\cdot)$ is a one to one mapping.

The **Fenchel-Young** inequality (3) is fundamental to convex analysis and plays an important role in our analysis.

$$\psi(\boldsymbol{y}) + \phi(\boldsymbol{x}) - \langle \boldsymbol{y}, \boldsymbol{x}\rangle \geq 0. \quad (3)$$

### 3.2 Properties

The convexity of (2) in $\boldsymbol{r}$ and $\boldsymbol{w}$ (separately) can be proven by verifying the identity (by evaluating its LHS):

$$D_\psi\big(\nabla\phi(\boldsymbol{y})\big|\big|\nabla\phi(\boldsymbol{x})\big) = D_\phi(\boldsymbol{x}||\boldsymbol{y}). \quad (4)$$

#### 3.2.1 Universality Of Minimizers

A mean-variance like decomposition (described in the appendix, Theorem (3)) holds for all Bregman divergences. It plays a critical role in Theorem 1 that has significant impact in facilitating the solution of the LETOR problem.

**Theorem 1.** *For $\mathcal{R}\downarrow \subset \mathbb{R}^d$ the entire set of vectors with descending ordered components, the minimizer $\boldsymbol{y}^* = \underset{\boldsymbol{y} \in \mathcal{R}\downarrow}{\operatorname{Argmin}} D_\phi(\boldsymbol{x}||\boldsymbol{y})$ is independent of $\phi(\cdot)$ if $\phi(\cdot)$ is WIS.*

*Proof: sketched in the appendix.*

Following our independent proof of Theorem 1, we have since come across an older proof [20] in the context of maximum likelihood estimators of exponential family models under conic constraints that were developed prior to the popularity of Bregman divergences. Whereas the older proof uses Moreau's cone decomposition[21], ours uses Theorem 3 (in appendix) and yields a much shorter proof.

**Corrolary 1.** *If* $\text{dom}\,\psi(\cdot) = \mathbb{R}^d$ *where* $\psi(\cdot)$ *is the Legendre conjugate of the WIS convex function* $\phi(\cdot)$ *then* $\text{Argmin}_{\boldsymbol{y} \in \mathcal{R}\downarrow \cap \text{dom}\,\phi}\, D_\phi\big(\boldsymbol{y}\big\|(\nabla\phi)^{-1}(\boldsymbol{x})\big) = (\nabla\phi)^{-1}(\boldsymbol{y}^*)$ *where* $\boldsymbol{y}^* = \text{Argmin}_{\boldsymbol{y} \in \mathcal{R}\downarrow}\,||\boldsymbol{x} - \boldsymbol{y}||^2$.

Corollary (1) implies that for choices of convex function $\phi(\cdot)$ indicated, the minimization over $\boldsymbol{r}_i \in \mathcal{R}\downarrow \cap \text{dom}\,\phi$ can be obtained by transforming the equivalent squared loss minimizer by $(\nabla\phi)^{-1}(\cdot)$. The squared loss minimization is not only simpler but its implementation can now be shared across all different $\phi(\cdot)$s where Corollary 1 applies. This class of convex functions is the same as "essentially smooth" [21]. Three such functions are listed in Table 1.

### 3.2.2 Optimality of Sorting

For any sorted vector $\boldsymbol{x}$, finding the permutation of $\boldsymbol{y}$ that minimizes $D_\phi(\boldsymbol{x}\|\boldsymbol{y})$ shows up as a subproblem in our formulation that needs to be solved in an inner loop. Thus solving it efficiently is critical and this is yet another instance where Bregman divergences are very useful.

For an arbitrary divergence function the search over the optimal permutation is a *non-linear assignment* problem that can be solved only by exhaustive enumeration. For an arbitrary separable divergence the optimal permutation may be found by solving a linear assignment problem, which is an integer linear program and hence also expensive to solve (especially in an inner loop, as required in our algorithm).

On the other hand, if $\phi(\cdot)$ is IS, the solution is remarkably simple, as shown in Lemma 3 where $\begin{bmatrix}\boldsymbol{x}_1\\\boldsymbol{x}_2\end{bmatrix}$ denotes a vector in $\mathbb{R}^2$ with components $x_1$ and $x_2$.

**Lemma 3.** *If* $x_1 \geq x_2$ *and* $y_1 \geq y_2$ *and* $\phi(\cdot)$ *is IS, then* $D_\phi(\begin{bmatrix}\boldsymbol{x}_1\\\boldsymbol{x}_2\end{bmatrix}\|\begin{bmatrix}\boldsymbol{y}_1\\\boldsymbol{y}_2\end{bmatrix}) \leq D_\phi(\begin{bmatrix}\boldsymbol{x}_1\\\boldsymbol{x}_2\end{bmatrix}\|\begin{bmatrix}\boldsymbol{y}_2\\\boldsymbol{y}_1\end{bmatrix})$ *and* $D_\phi(\begin{bmatrix}\boldsymbol{y}_1\\\boldsymbol{y}_2\end{bmatrix}\|\begin{bmatrix}\boldsymbol{x}_1\\\boldsymbol{x}_2\end{bmatrix}) \leq D_\phi(\begin{bmatrix}\boldsymbol{y}_2\\\boldsymbol{y}_1\end{bmatrix}\|\begin{bmatrix}\boldsymbol{x}_1\\\boldsymbol{x}_2\end{bmatrix})$

*Proof.* $D_\phi(\begin{bmatrix}\boldsymbol{x}_1\\\boldsymbol{x}_2\end{bmatrix}\|\begin{bmatrix}\boldsymbol{y}_1\\\boldsymbol{y}_2\end{bmatrix}) - D_\phi(\begin{bmatrix}\boldsymbol{x}_1\\\boldsymbol{x}_2\end{bmatrix}\|\begin{bmatrix}\boldsymbol{y}_2\\\boldsymbol{y}_1\end{bmatrix}) = \langle(\nabla\phi(y_2) - \nabla\phi(y_1)), x_1 - x_2\rangle$. There exists $c \geq 0$ s.t. $x_1 - x_2 = c(y_1 - y_2)$. Proof follows from monotonicity of $\nabla\phi$, ensured by convexity of $\phi$. We can exchange the order of the arguments using (4). □

Using induction over $d$ for $\boldsymbol{y} \in \mathbb{R}^d$ the optimal permutation is obtained by sorting. Not only is Lemma 3 extremely helpful in generating descent updates, it has fundamental consequences in relation to the local and global optimum of our formulation (see Lemma 4).

### 3.2.3 Joint Convexity and Global Minimum

Using Legendre duality one recognizes that equation (2) quantifies the gap in the Fenchel-Young inequality (3).

$$D_\phi\big(\boldsymbol{r}_i\big\|(\nabla\phi)^{-1}(\boldsymbol{A}_i\boldsymbol{w})\big) = \psi(\boldsymbol{A}_i\boldsymbol{w}) + \phi(\boldsymbol{r}_i) - \langle\boldsymbol{r}_i, \boldsymbol{A}_i\boldsymbol{w}\rangle.$$

Although this clarifies the issue of separate convexity in $\boldsymbol{w}$ and $\boldsymbol{r}_i$, the conditions under which joint convexity is obtained is not obvious. Joint convexity, if ensured, guarantees global minimum even for a coordinate-wise minimization because our constraint set is a product of convex sets. We resolve this important question in Theorem 2.

**Theorem 2.** *The gap in the Fenchel-Young inequality* $\psi(\boldsymbol{y}) + \phi(\boldsymbol{x}) - \langle\boldsymbol{x}, \boldsymbol{w}\rangle$ *for any twice differentiable strictly convex* $\phi(\cdot)$ *with a differentiable conjugate* $(\phi)^*(\cdot) = \psi(\cdot)$ *is jointly convex if and only if* $\phi(\boldsymbol{x}) = c||\boldsymbol{x}||^2$ *for all* $c > 0$.

*Proof: sketched in the appendix.*

### 3.3 Algorithms

Now we discuss Bregman's algorithm associated with Bregman divergences. The original motivation for introducing [5] Bregman divergence was to generalize alternating orthogonal projection. A significant advantage of the algorithm is its scalability and suitability for parallelization. It solves the following (Bregman projection) problem:

$$\min_{\boldsymbol{x}} D_\phi(\boldsymbol{x}\|\boldsymbol{y}) \text{ s.t. } \boldsymbol{A}\boldsymbol{x} \leq \boldsymbol{b} \qquad (5)$$

---

**Bregman's algorithm:**

**Initialize:** $\boldsymbol{\lambda}^0 \in \mathbb{R}+^d$ and $z^0$ such that
$\nabla\phi(z^0) = \big[\boldsymbol{A}^\dagger|\nabla\phi(\boldsymbol{y})\big]\big[\boldsymbol{\lambda}^{0\dagger}, 1\big]^\dagger$

**Repeat:** Till convergence

    **Update:** Apply **Sequential** or **Parallel Update**

    **Solve:** $\nabla\phi(z^{t+1}) = \big[\boldsymbol{A}^\dagger|\nabla\phi(\boldsymbol{y})\big]\big[\boldsymbol{\lambda}^{t+1\,\dagger}, 1\big]^\dagger$

---

**Sequential Bregman Update:**

**Select** $i$: Let $\mathcal{H}_i = \{\boldsymbol{z}|\,\langle\boldsymbol{a}_i, \boldsymbol{z}\rangle \leq b_i\}$

**If in violation:** Compute $\text{Proj}^\phi\,(\boldsymbol{z}^t, \mathcal{H}_i)$ i.e.
$\nabla\phi(\text{Proj}^\phi\,(\boldsymbol{z}^t, \mathcal{H}_i)) = \nabla\phi(\boldsymbol{z}^t) + c_i^t\boldsymbol{a}_i,$

**Update:** $\boldsymbol{\lambda}^{t+1} = \boldsymbol{\lambda}^t + c_i^t\boldsymbol{1}_i$

---

**Parallel Bregman Update:**

**For all** $i$ **in parallel:** Compute $\text{Proj}^\phi\,(\boldsymbol{z}^t, \mathcal{H}_i)\,, c_i^t$

**Update:** $\boldsymbol{\lambda}_i^{t+1} = \boldsymbol{\lambda}^t + c_i^t\boldsymbol{1}_i$

**Synchronize:** $\boldsymbol{\lambda}^{t+1} = \nabla^{-1}\phi(\sum_i \nabla\phi(\boldsymbol{\lambda_i}^{t+1}))$

---

## 4 LETOR with Monotone Retargeting

Our cost function is an instantiation of (2) with a WIS Bregman divergence. In addition, we include regularization and a query specific offset. Note that the cost function (2) is not invariant to scale. For example squared Euclidean, KL divergence and generalized I-divergence are homogeneous functions of degree 2, 1 and 1 respectively. Thus the cost can be reduced just by scaling its arguments down, without actually learning the task. To remedy this, we restrict the $r_i$s from shrinking below a pre-prescribed size. This is accomplished by constraining $r_i$s to lie in an appropriate closed convex set separated from the origin, for example, an unit simplex or a shifted positive orthant. This yields:

$$\min_{\beta_i, w, r_i \in \mathcal{R}\downarrow_i \cap \mathcal{S}_i} \sum_{i=1}^{|\mathcal{Q}|} \frac{1}{|\mathcal{V}_i|} D_\phi \big(r_i \big\| (\nabla\phi)^{-1} \left(A_i w + \beta_i \mathbf{1}\right)\big) + \frac{C}{2}\|w\|^2, \quad (6)$$

or equivalently

$$\min_{\beta_i, w, r_i \in \mathcal{R}\downarrow_i \cap \mathcal{S}_i} \sum_{i=1}^{|\mathcal{Q}|} \frac{1}{|\mathcal{V}_i|} D_\psi \big(A_i w + \beta_i \mathbf{1} \big\| \nabla\phi\left(r_i\right)\big) + \frac{C}{2}\|w\|^2, \quad (7)$$

where $\mathcal{S}_i$ are bounded sets excluding $\mathbf{0}$, chosen to suit the divergence. The parameter $C$ is the regularization parameter. In non-transductive settings, the query specific offsets $\beta_i$ will not be available for the test queries. This causes no difficulty because $\beta_i$ does not affect the relative ranks over the documents. We update the $r_i$'s and $\{w, \{\beta_i\}\}$ alternately. Note that each is a Bregman projection.

If $\mathcal{S}_i = \text{dom}\,\phi$ and $\text{dom}\,\psi = \mathbb{R}^d$, the optimization over $r_i$ reduces to an order constrained least squares problem (corrolary-1). Examples of such matched pairs are (i) $wKL\left(\cdot\|\cdot\right)$ and $\Delta_i$, and (ii) shifted $wGI\left(\cdot\|\cdot\right)$ and $\mathbf{1}+\mathbb{R}_+{}^d$. A well studied, scalable algorithm for the ordered least squares problem is pool of adjacent violators (PAV) algorithm [3]. One can verify that PAV, like Bregman's algorithm (5) is a dual feasible method. One may also use Lemma 1 to solve it as a non-negative least squares problem for which several scalable algorithm exists [15].

To be able to use Bregman's algorithm it is essential that $\mathcal{R}\downarrow_i$ be available as an intersection of linear constraints, as is readily obtained for any prescribed total order, as shown:

$$\mathcal{R}\downarrow_i = \{r_{i,j+1} - r_{i,j} \leq 0\}_{\forall j \in \mathcal{J}_i},$$
$$\Delta_i^o = \mathcal{R}\downarrow_i \cap \{\sum_j r_{ij} = 1\} \cap \{r_{i,d_i} > 0\}. \quad (8)$$

Partial orders are discussed in section 4.1.

The advantages of the Bregman updates (3.3), are that they are easy to implement (more so when $\text{Proj}^\phi\left(\cdot,\cdot\right)$ is available in closed form e.g. squared Euclidean), have minimal memory requirements, and hence they scale readily and allow easy switch from a sequential to a parallel update.

The parallel Bregman updates applied to (2), (8) clearly exposes massive amounts of fine grained parallelism at the level of individual inequalities in $\mathcal{R}\downarrow_i$ or $\Delta_i^o$, and is well suited for implementation on a GPGPU[18]. We note further that the optimization for $r_i$ is independent for each query, thus can be embarrassingly parallelized.

For optimizing over $w$ one may use several techniques available for parallelizing a sum of convex functions, for example parallelize the gradient computation across the terms or use more specialized technique such as alternating direction of multipliers [4]. Further, $\{w, \{\beta_i\}\}$ can be solved jointly simply by augmenting the feature matrix $A_i$ with $\mathbf{1}$. We hope the readers will appreciate this flexibility of being able to exploit parallelism at different levels of granularity of choice.

### 4.1 Partial Order

Recall that a partial order is induced if the number of unique rank scores $k_i$ in $\tilde{r}_i$ is less than $d_i$. In this case our convention of indexing $\mathcal{V}_i$ in a descending order is ambiguous. To resolve this, we break ties arbitrarily. Consider a subset of $\mathcal{V}_i$ whose elements have the same training rank-score. We distinguish between two modeling choices: (a) the items in that subset are not really equivalent, but the training set used a resolution that could not make fine distinctions between the items,[2] we call this the "hidden order" case, and (b) the items in the subset are indeed equivalent and the targets are constrained to reflect the same block structure, we call this case "block equivalent" and can model it appropriately. Although we have removed the discussion on the latter in the interest of space, this too can be modeled efficiently by MR.

#### 4.1.1 Partially Hidden Order

In this model we assume that the items are totally ordered, though the finer ordering between similar items is not visible to the ranking algorithm. Let $P_i = \{P_{ik}\}_{k=1}^{k_i}$ be a partition of the index set of $\mathcal{V}_i$, such that all items in $P_{ik}$ have the same training rank-score. We denote their sizes by $d_{ik} = |P_{ik}|$. The sets $\mathcal{V}_i$ effectively get partitioned further into $\{P_{ik}\}_{k=1}^{k_i}$ by the $k_i < d_i$ unique scores given to each of its members. Though such a score specifies an order between items from any two different sets $P_{ij}$ and $P_{il}$, the order within any set $P_{ik}$ remains unknown. This is very common in practice and is usually an artifact of the high cost of acquiring training data in a totally ordered form. The optimization problem may be solved using either an inner or an outer representation of the constraint sets.

**Outer representation:** Recall that Bregman's algorithm 3.3 is better suited for the outer representation (8).

Denote the set of rank-score vectors having the same partially ordered structure as $\tilde{r}_i$ by $\mathfrak{R}_i$. For partial order we may describe $\mathfrak{R}_i$ by linear inequalities as follows:

$$\{r_{im} > r_{in}\}_{j=1}^{k_i-1} \forall_{i \in [1, |\mathcal{Q}|]}, \; m \in P_{ij}, \; n \in P_{i,j+1},$$

with each $j$ generating $d_{ij} d_{i,j+1}$ inequalities, which is very high. The proliferation of inequalities may be reduced by

$$\boldsymbol{x}_i^{t+1} = \underset{\boldsymbol{x} \in \Delta}{\text{Argmin}} \, D_\phi(T\boldsymbol{x} \big\| (\nabla \phi)^{-1} \left( \mathbb{P}_i^t \boldsymbol{A}_i \boldsymbol{w}^t + \beta_i^t \right)) \quad \forall i \quad (10)$$

$$\mathbb{P}_i^{t+1} = \underset{\pi}{\text{Argmin}} \, D_\phi(T\boldsymbol{x}_i^{t+1} \big\| (\nabla \phi)^{-1} \left( \pi \boldsymbol{A}_i \boldsymbol{w}^t + \beta_i^t \right)) \quad \forall i \quad (11)$$

$$\boldsymbol{w}^{t+1}, \{\beta_i^{t+1}\} = \quad (12)$$
$$\underset{\boldsymbol{w}, \{\beta_i\}}{\text{Argmin}} \sum_{i=1}^{|\mathcal{Q}|} D_\phi(T\boldsymbol{x}^{t_i+1} \big\| (\nabla \phi)^{-1} \left( \mathbb{P}_i^{t+1} \boldsymbol{A}_i \boldsymbol{w} + \beta_i^t \right)) + \frac{C}{2} \|\boldsymbol{w}\|^2$$

---

[2]or that, we only care to reduce the error of predicting $r_{ij} > r_{ik}$ when $\tilde{r}_{ij} < \tilde{r}_{ik}$, note the strict inequality.

Figure 1: Algorithm for Partially Hidden Order

introducing auxiliary variables $\{\bar{r}_{i,l}\}_{l=1}^{k_i-1}$ and the following inequalities:

$$\{\bar{r}_{i,j+1} > r_{iP_{ij}} > \bar{r}_{i,j}\} \forall_{i \in [1,|\mathcal{Q}|]}. \quad (9)$$

However, since Bregman's algorithms are essentially coordinate-wise ascent methods, their convergence may slow unless fine grained parallelism can be exploited. For commodity hardware, an alternative to the exterior point methods are interior point methods that use an inner representation of the convex constraint set.

**Inner representation:** For our experiments we use the updates in figure 1. In particular, we use the method of $D$ proximal gradients for (10) where the proximal term is a Bregman divergence defined by a convex function whose domain is the required constraint set [13], [22], [16]. This automatically enforces the required constraints.

To handle partial order we introduce a block-diagonally restricted permutation matrix $\mathbb{P}_i$ that can permute indices in each $P_{ij}$ independently. Since the items in $P_{ij}$ are not equivalent they are available for re-ordering as long as that minimizes the cost (6). Block weighted IS Bregman divergences have the special property that sorting minimizes the divergence over all permutations (Lemma 3). Thus update (11) can be accomplished by sorting.

The updates (10), (11) and (12) each reduce the lower bounded cost (6), and therefore the algorithm described in figure 1 converges. However, the vital question about whether the updates converge to a stationary point remains.

**Convergence to a Stationary Point** If repeated application of (10) and (11) (sorting) for a fixed $\boldsymbol{w}^{t+1}, \{\beta_i^{t+1}\}$ achieves the minimum then convergence to the stationary point is guaranteed. Thus we explore the question whether (11) and (10) together achieves a local minimum.

The tri-factored form $\boldsymbol{r}_i \mathbb{P}_i U \boldsymbol{x}_i$ is a cause for concern. Somewhat re-assuring is the fact that the range of $\mathbb{P}_i U \boldsymbol{x}_i$ is $\mathfrak{R}_i$ which again is a convex cone and that the tri-factored representation of any point in that cone is described uniquely. This however is not sufficient to ensure that a minimum is achieved by (10) and (11) because though the constraint set is convex, the cost function (6) is not convex in the tri-factored parameterization. Worse still, the parameterization is discontinuous because of the discrete nature of $\mathbb{P}$.

While one may address the discreteness problem via a real-relaxation of $\mathbb{P}$ to doubly stochastic matrices, the local minima attained in such a case will be in the interior of the Birkhoff polytope and not at the vertices that (11) sorting would have obtained. Therefore such a convex relaxation cannot answer the question whether (10) and sorting achieves the local minimum. Thus it is surprising that sorting followed by the $\boldsymbol{x}_i$ updates does achieve the local minimum of (6) on the cone $\mathfrak{R}_i$, as a consequence of the following Lemma.

**Lemma 4.** *Let vectors $\begin{bmatrix} \boldsymbol{x}_1 \\ \boldsymbol{x}_2 \end{bmatrix}$ and $\begin{bmatrix} \boldsymbol{y}_1 \\ \boldsymbol{y}_2 \end{bmatrix}$ be conformally partitioned. Let $\begin{bmatrix} \boldsymbol{y}_1 \\ \boldsymbol{y}_2 \end{bmatrix}^* = \underset{\substack{\boldsymbol{y}_i' \in \Pi(\boldsymbol{y}_1), \\ \boldsymbol{y}_1' \geq \boldsymbol{y}_2'}}{\text{Argmin}} \, D_\phi\left(\begin{bmatrix} \boldsymbol{y}'_1 \\ \boldsymbol{y}'_2 \end{bmatrix} \big\| \begin{bmatrix} \boldsymbol{x}_1 \\ \boldsymbol{x}_2 \end{bmatrix}\right)$ where $\Pi(\boldsymbol{y}_i)$ is the set of all permutations of the vector $\boldsymbol{y}_i$. If the Bregman divergence $D_\phi(\cdot \| \cdot)$ is conformally separable then $\boldsymbol{y}_i^*$ is isotonic with $\boldsymbol{x}_i \, \forall i = 1, 2$*

*Proof.* The proof is by contradiction. Assume $\boldsymbol{y}_i^*$ is a minimizer that is not isotonic with $\boldsymbol{x}_i$, then one may permute $\boldsymbol{y}_i^*$ to match the order of $\boldsymbol{x}_i$ to obtain a reduced cost, yielding a contradiction. $\square$

The utility of Lemma 4 is that in spite of the caveats mentioned it can correctly identify the internal ordering of the components of the left hand side that achieves the minimum for a fixed $\boldsymbol{w}^{t+1}, \{\beta_i^{t+1}\}$, given a fixed right hand side. With the knowledge of the order obtained, one may then compute the actual values with relative ease with (10).

## 5 Experiments

We evaluated the ranking performance of the proposed monotone retargeting approach on the benchmark LETOR 4.0 datasets (MQ2007, MQ2008) [23] as well as the OHSUMED dataset [12]. Each of these datasets is pre-partitioned into five-fold validation sets for easy comparison across algorithms. For OHSUMED, we used the *QueryLevelNorm* partition. Each dataset contains a set of queries, where each document is assigned a relevance score from irrelevant ($r = 0$) to relevant ($r = 2$).

All algorithms were trained using a regularized linear ranking function, with a regularization parameter chosen from the set $C \in \{10^{-50}, 10^{-20}, 10^{-10}, 10^{-5}, 10^0, 10^1\}$. The best model was identified as the model with highest mean average precision (MAP) on the validation set. All presented results are of average performance on the test set. As the baseline, we implemented the NDCG consistent renormalization approach in [19] (using the NDCG$_m$ normalization) for the squared loss and the I-divergence (generalized KL-divergence). ListNet was implemented [7] as the KL divergence baseline since their normalization has no effect on KL-divergence. MR was implemented using the *partially hidden order* monotone retargeting approach (Section 4.1). We compared the performance of MR (Normalized MR) to the MR method with the normalization $\frac{1}{|\mathcal{V}_i|}$ removed (Unnormalized MR).

The algorithms were implemented in Python and executed on a 2.4GHz quad-core Intel Xeon processor without paying particular attention to writing optimized code. Ample room for improvement remains. Square loss was the fastest with respect to average execution times per iteration at 0.58 seconds whereas KL achieved 1.01 seconds per iteration and I-div 1.14 seconds per iteration. We found that although MQ2007 is more than 4 times larger than MQ2008, MQ2007 only required about twice the time execution on average, highlighting the scalability of MR. On average SQ, KL and I-div took 99, 90 and 65 iterations.

Table 5 compares the algorithms in terms of expected reciprocal return (ERR) [10], Mean average precision (MAP) and NDCG. Unnormalized KL divergence cost function led to the best performance across datasets. The most significant gains over the baseline were for the I-divergence cost function. Monotone retargeting showed consistent performance gains over the baseline across metrics (NDCG, ERR, Precision), suggesting the effectiveness of MR for improving the overall ranking performance.

Figure 2 shows a subset of performance comparisons using the NDCG@N and Precision@N metrics. Our experiments show a significant improvement in performance on the range of datasets and cost functions. Across datasets, the difference between the baseline and our results were most significant with the I-divergence (generalized KL divergence) cost function.

There are two things worth taking special note of: (i) Although the baseline algorithms were proposed specifically for improving NDCG performance, MR improves the ranking accuracy further, even in terms of NDCG. (ii) MR seems to be consistently peaking early. This property is particularly desirable and is encoded specifically in the cost functions such as NDCG and ERR. In our initial formulation we used WIS Bregman divergence so that the weights could be tuned to obtained the early peaking behavior. However that proved unnecessary because even the unweighted model produced satisfactory performance. The effect of query length normalization was, however, inconsistent. Some of our results were insensitive to it, whereas other results were adversely affected. We conjecture that it is an artifact of using the same amount of regularization as in the un-normalized case.

## 6 Conclusion and Related Work

One technique that shares our motivation of learning to rank is ordinal regression [14] which optimizes parameters of a regression function as well as thresholds. Unfortunately we do not have space for a full literature survey and only mention a few key differences. The log-likelihood of the classic ordinal regression methods are a sum of logarithms of differences of monotonic functions and are much more cumbersome to optimize over. To our knowledge they do not share the strong guarantees that MR with Bregman divergences enjoys. The prevalent technique there seems to require fixing a finite number of thresholds up-front, an arbitrary choice that MR does not make. However, it is not clear if that is a restriction of ordinal regression or is a prevalent practice.

| MQ 2007 ERR | | | |
|---|---|---|---|
| | I-div | SQ | KL |
| Unnormalized MR | 0.3698 | 0.3703 | **0.3737** |
| Normalized MR | 0.3702 | 0.3601 | 0.3731 |
| Baseline | 0.1953 | 0.3639 | 0.3643 |
| MQ 2007 MAP | | | |
| | I-div | SQ | KL |
| Unnormalized MR | 0.5379 | 0.5361 | 0.5398 |
| Normalized MR | 0.5358 | 0.5282 | **0.5399** |
| Baseline | 0.3611 | 0.5330 | 0.5380 |
| MQ 2007 NDCG | | | |
| | I-div | SQ | KL |
| Unnormalized MR | 0.6961 | **0.7398** | 0.6978 |
| Normalized MR | 0.6954 | 0.6953 | 0.6981 |
| Baseline | 0.5512 | 0.6927 | 0.6952 |
| MQ 2008 ERR | | | |
| | I-div | SQ | KL |
| Unnormalized MR | 0.4137 | 0.41559 | **0.4238** |
| Normalized MR | 0.4144 | 0.41392 | 0.4085 |
| Baseline | 0.2724 | 0.40978 | 0.4132 |
| MQ 2008 MAP | | | |
| | I-div | SQ | KL |
| Unnormalized MR | 0.6439 | 0.6532 | **0.6571** |
| Normalized MR | 0.6449 | 0.6549 | 0.6461 |
| Baseline | 0.4513 | 0.6428 | 0.6530 |
| MQ 2008 NDCG | | | |
| | I-div | SQ | KL |
| Unnormalized MR | 0.7339 | 0.7398 | **0.7451** |
| Normalized MR | 0.7346 | 0.7396 | 0.7330 |
| Baseline | 0.5892 | 0.7344 | 0.7399 |
| OHSUMED ERR | | | |
| | I-div | SQ | KL |
| Unnormalized MR | 0.5657 | 0.5410 | 0.5410 |
| Normalized MR | **0.5796** | 0.5093 | 0.5093 |
| Baseline | 0.2255 | 0.5450 | 0.5467 |
| OHSUMED MAP | | | |
| | I-div | SQ | KL |
| Unnormalized MR | **0.4537** | 0.4417 | 0.4531 |
| Normalized MR | 0.4463 | 0.4394 | 0.4506 |
| Baseline | 0.3421 | 0.4465 | 0.4524 |
| OHSUMED NDCG | | | |
| | I-div | SQ | KL |
| Unnormalized MR | **0.7000** | 0.6878 | 0.6997 |
| Normalized MR | 0.6935 | 0.6798 | 0.6916 |
| Baseline | 0.5805 | 0.6892 | 0.6947 |

Table 2: Test ERR, MAP and NDCG on different datasets. The best results are in bold.

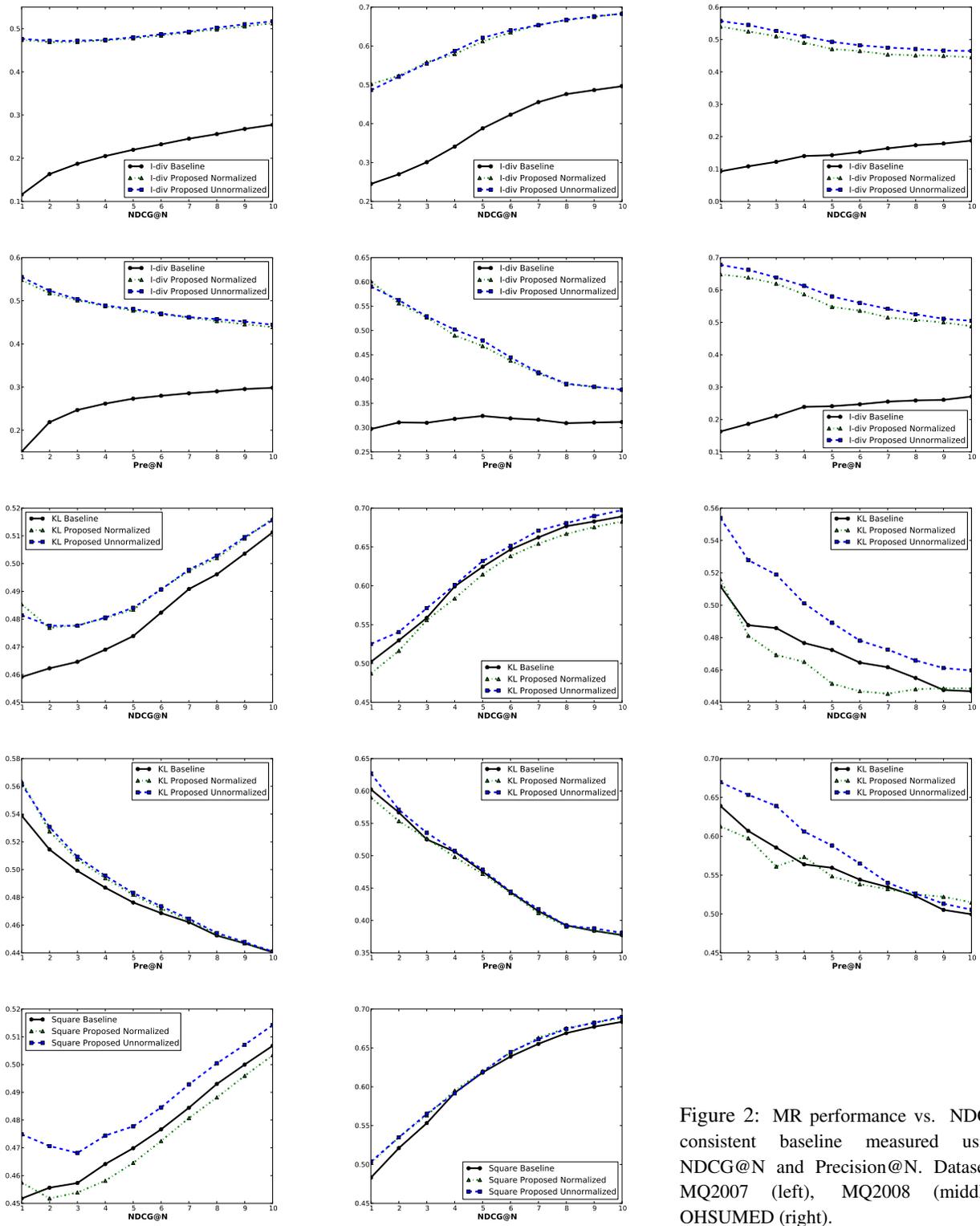

Figure 2: MR performance vs. NDCG consistent baseline measured using NDCG@N and Precision@N. Datasets: MQ2007 (left), MQ2008 (middle), OHSUMED (right).

In this paper we introduced a family of new cost functions for ranking. The cost function takes into account all possible monotonic transforms of the target scores, and we show how such a cost function can be optimized efficiently. Because the sole objective of learning to rank is to output good permutations on unseen data, it is desirable that the cost function be a function of such permutations. Though several permutation dependent cost functions have been proposed, they are extremely difficult to optimize over and one has to resort to surrogates and/or cut other corners. We show that with monotone retargeting with Bregman divergences such contortions are unnecessary. In addition, the proposed cost function and algorithms have very favorable statistical, optimization theoretic, as well as empirically observed properties. Other advantages include extensive parallelizability due to simple simultaneous projection updates that optimize a cost function that is convex not only in each of the arguments separately but also jointly, with a proper choice of the cost function from the family.

## A  Appendix: Proof Sketches

**Theorem 1**

*Proof.* Let the components of $\boldsymbol{y}^*$ take $k$ unique values. Partition the set indexing the components into $\Pi = \{\Pi_i\}_{i=1}^k$ s.t. $\forall_j \in \Pi_i \; y_j^* = c_i \; \forall_{i \in [1,k]}$. Let the scalar mean of $\boldsymbol{x}$ on $\Pi_i$ be $\mu_{\Pi_i}$. By (14), $\sum_{j \in \Pi_i} D_\phi(x_j \| y_j^*) = \sum_{j \in \Pi_i} D_\phi(x_j \| \mu_{\Pi_i}) + D_\phi(\mu_{\Pi_i} \| c_i)$. First, we prove by contradiction that $y_j^* = \mu_{\Pi_i} \; \forall_j \in \Pi_i$, otherwise $\exists \; \boldsymbol{y}' \in \mathbb{R}^d$ s.t. $y_l' = y_l^* \; \forall l \notin \Pi_i$, and $\forall_j \in \Pi_i \; c_{i+1} \leq y_j' = c' \leq c_{i-1}$ s.t. $D_\phi(\mu_{\Pi_i} \| c') < D_\phi(\mu_{\Pi_i} \| c_i)$. Thus $D_\phi(\boldsymbol{x} \| \boldsymbol{y}') < D_\phi(\boldsymbol{x} \| \boldsymbol{y}^*)$, clearly a contradiction.

Let $\operatorname{Argmin}_{\boldsymbol{y} \in \mathcal{R}\downarrow} D_\theta(\boldsymbol{x} \| \boldsymbol{y}) = \boldsymbol{z}^*$ for $\boldsymbol{\theta} \neq \boldsymbol{\phi}$. Let $\boldsymbol{z}^*$ induce the partition $P = \{P_l\}_{l=1}^m$. If $\Pi = P$ always, then $y_j^* = z_j^*$ completing the proof. Shift the indexing of the partitions to the first $j$ where $\Pi_j, P_j$ differs. Now with new index, WLOG [3] assume $\Pi_1 \subset P_1$, $P_1 \subset \Pi_1 \cup \Pi_2$. Define $\dot{\Pi}_2 = \Pi_2 \cap P_1, \ddot{\Pi}_2 = \Pi_2 \setminus \dot{\Pi}_2 \neq \emptyset$, else $P_1$ can be refined

---

[3]We encourage the reader to draw a picture for clarity.

into $\Pi_1, \Pi_2$ obtaining a lower cost (by Corollary 3). Let the means of $\dot{\Pi}_2, \ddot{\Pi}_2 = \dot{y}_2, \ddot{y}_2$. By definition $y_2^*$ is their convex combination. Now $\dot{y}_2 \geq \ddot{y}_2$, else one can reduce the cost by refining $\Pi_2$ into $\dot{\Pi}_2, \ddot{\Pi}_2$. Therefore $\dot{y}_2 \geq y_2^* \geq \ddot{y}_2$. Note $y_1^* \geq \dot{y}_2$ or else we can refine $P_1$ into $\Pi_1, \dot{\Pi}_2$. Thus we have $y_1^* \geq \dot{y}_2 \geq \ddot{y}_2$ which is in contradiction with $y_1^* < y_2^*$. □

**Theorem 2**

*Proof.* For succinctness we use the abbreviations: $\boldsymbol{x}(\alpha) = \alpha \boldsymbol{x}_1 + (1-\alpha)\boldsymbol{x}_2$, $\boldsymbol{y}(\alpha) = \alpha \boldsymbol{y}_1 + (1-\alpha)\boldsymbol{y}_2$, $\phi_i = \phi(\boldsymbol{x}_i)$, $\psi_i = \psi(\boldsymbol{x}_i)$, $\Phi(\alpha) = \alpha\phi_1 + (1-\alpha)\phi_2$ and $\Psi(\alpha) = \alpha\psi_1 + (1-\alpha)\psi_2$. Joint convexity is equivalent to $\phi(\boldsymbol{x}(\alpha)) + \psi(\boldsymbol{y}(\alpha)) - \langle \boldsymbol{x}(\alpha), \boldsymbol{y}(\alpha) \rangle \leq \Phi(\alpha) + \Psi(\alpha) - \alpha \langle \boldsymbol{x}_1, \boldsymbol{y}_1 \rangle - (1-\alpha) \langle \boldsymbol{x}_2, \boldsymbol{y}_2 \rangle \; \forall \; \boldsymbol{x}_1, \boldsymbol{x}_2 \in \operatorname{dom} \phi$, $\boldsymbol{y}_1, \boldsymbol{y}_2 \in \operatorname{dom} \psi$. Thus we have to show:

$$\phi(\boldsymbol{x}(\alpha)) + \psi(\boldsymbol{y}(\alpha)) \leq \Phi(\alpha) + \Psi(\alpha) + \overbrace{\alpha(1-\alpha) \langle \boldsymbol{x}_1 - \boldsymbol{x}_2, \boldsymbol{y}_1 - \boldsymbol{y}_2 \rangle}^{B} \quad (13)$$

for all arguments in the domain. Assume with no loss in generality that $\phi(\cdot)$ and $\psi(\cdot)$ are strongly convex with modulus of strong convexity $(1 + s1), (1 - s2)$ with $s1 > -1, s2 < 1$, respectively. Reciprocal of the modulus of strong convexity of the Legendre dual is the Lipschitz constant of the gradient of a convex function [21], therefore $(1 + s1) \leq \frac{1}{1-s2}$, being the lower and upper bounds of the eigenvalues of the Hessian of $\phi(\cdot)$ respectively. Simplifying expression (13) using our strong convexity assumptions and positivity of $\alpha(1-\alpha)$, we obtain that we have to show $(1 + s1)\|\boldsymbol{x}_1 - \boldsymbol{x}_2\|^2 + (1 - s2)\|\boldsymbol{y}_1 - \boldsymbol{y}_2\|^2 - 2B \leq 0$. Or

$$\|(\boldsymbol{x}_1 - \boldsymbol{x}_2) - (\boldsymbol{y}_1 - \boldsymbol{y}_2)\|^2 + s1\|\boldsymbol{x}_1 - \boldsymbol{x}_2\|^2 - s2\|(\boldsymbol{y}_1 - \boldsymbol{y}_2)\|^2 \leq 0.$$

Let $\boldsymbol{p} = \boldsymbol{x}_1 - \boldsymbol{x}_2$ and $\boldsymbol{q} = \boldsymbol{y}_1 - \boldsymbol{y}_2$. By choosing $(1+s)\boldsymbol{p} = \boldsymbol{q}$ we obtain $s1 > s2 + s1s2$, or equivalently $(1 + s1) \geq \frac{1}{1-s2}$. Thus the lower and upper bounds of the eigenvalues of Hessian of $\phi(\cdot)$ must coincide. □

## B  Appendix: Optimality of Means

**Theorem 3.** *[2] Let $\boldsymbol{\pi}$ be a distribution over $\boldsymbol{x} \in \operatorname{dom}\phi$ and $\boldsymbol{\mu} = \mathbb{E}_{\boldsymbol{x} \sim \boldsymbol{\pi}}[\boldsymbol{x}]$ then the expected divergence about $\boldsymbol{s}$ is*

$$\mathbb{E}_{\boldsymbol{x} \sim \boldsymbol{\pi}}\left[D_\phi(\boldsymbol{x}\|\boldsymbol{s})\right] = \mathbb{E}_{\boldsymbol{x} \sim \boldsymbol{\pi}}\left[D_\phi(\boldsymbol{x}\|\boldsymbol{\mu})\right] + D_\phi(\boldsymbol{\mu}\|\boldsymbol{s}). \quad (14)$$

From non-negativity of Bregman divergence it follows that

**Corollary 2.** *[2]* $\mathbb{E}_{\boldsymbol{x} \sim \boldsymbol{\pi}}[\boldsymbol{x}] = \operatorname*{Argmin}_{\boldsymbol{y} \in \operatorname{dom}\phi} \mathbb{E}_{\boldsymbol{x} \sim \boldsymbol{\pi}}\left[D_\phi(\boldsymbol{x}\|\boldsymbol{y})\right].$

**Corollary 3.** *If random variable $\boldsymbol{x}$ takes values in $\mathcal{X} = \mathcal{X}_1 \cup \mathcal{X}_2$ with $\mathcal{X}_1 \cap \mathcal{X}_2 = \emptyset$ then* $\operatorname*{Argmin}_{\boldsymbol{\mu} \in \mathcal{X}} \mathbb{E}_{\boldsymbol{x}|\mathcal{X}}\left[D_\phi(\boldsymbol{x}\|\boldsymbol{\mu})\right] \geq$

$\operatorname*{Argmin}_{\boldsymbol{\mu}_1 \in \mathcal{X}_1} \mathbb{E}_{\boldsymbol{x}|\mathcal{X}_1}\left[D_\phi(\boldsymbol{x}\|\boldsymbol{\mu}_1)\right] + \operatorname*{Argmin}_{\boldsymbol{\mu}_2 \in \mathcal{X}_2} \mathbb{E}_{\boldsymbol{x}|\mathcal{X}_2}\left[D_\phi(\boldsymbol{x}\|\boldsymbol{\mu}_2)\right].$

### Acknowledgements

Authors acknowledge support from NSF grant IIS 1016614 and thank Cheng H. Lee for suggesting improvements over our initial submission.